\documentclass[conference, onecolumn]{IEEEtran}
\IEEEoverridecommandlockouts
\usepackage{balance}
\usepackage{amsmath,amssymb,amsfonts}
\usepackage{algorithmic}
\usepackage{booktabs}
\usepackage{cleveref}
\usepackage{graphicx}
\usepackage{textcomp}
\usepackage{xcolor}
\usepackage{biblatex}
\usepackage{xtab,booktabs}
\usepackage{tabularx}
\usepackage{caption}
\usepackage{subcaption}

\usepackage{color,soul}

\graphicspath{ {./images/} }

\def\BibTeX{{\rm B\kern-.05em{\sc i\kern-.025em b}\kern-.08em
    T\kern-.1667em\lower.7ex\hbox{E}\kern-.125emX}}
\begin{document}

\title{Integrating Heuristics and Learning in a Computational Architecture for Cognitive Trading
% {\footnotesize \textsuperscript{*}Note: Sub-titles are not captured in Xplore and
% should not be used}
% \thanks{Identify applicable funding agency here. If none, delete this.}
}

\author{
\IEEEauthorblockN{Remo Pareschi}
\IEEEauthorblockA{\textit{Stake Lab} \\
\textit{University of Molise}\\
remo.pareschi@unimol.it}
\and
\IEEEauthorblockN{Federico Zappone}
\IEEEauthorblockA{\textit{Stake Lab} \\
\textit{University of Molise}\\
f.zappone1@studenti.unimol.it}
}

\maketitle

\begin{abstract}
The successes of Artificial Intelligence in recent years in areas such as image analysis, natural language understanding and strategy games have sparked interest from the world of finance. Specifically, there are high expectations, and ongoing engineering projects, regarding the creation of artificial agents, known as robotic traders, capable of juggling the financial markets with the skill of experienced human traders. Obvious economic implications aside, this is certainly an area of great scientific interest, due to the challenges that such a real context poses to the use of AI techniques. Precisely for this reason, we must be aware that artificial agents capable of operating at such levels are not just round the corner, and that there will be no simple answers, but rather a concurrence of various technologies and methods to the success of the effort. In the course of this article, we review the issues inherent in the design of effective robotic traders as well as the consequently applicable solutions, having in view the general objective of bringing the current state of the art of robo-trading up to the next level of intelligence, which we refer to as Cognitive Trading. Key to our approach is the joining of two methodological and technological directions which, although both deeply rooted in the disciplinary field of artificial intelligence, have so far gone their separate ways: heuristics and learning.
\end{abstract}


\section{Introduction}
Trading is a particularly interesting area, as well as to date a relatively unexplored one, for artificial intelligence (AI), both for its theoretical implications and for its practical applicability. In fact, it shares some of the characteristics of strategy games, an area in which AI has achieved some of its most striking successes, yet it has to face open and unpredictable situations that have no place in the closed and stable universe of games. Its implementation through highly performing artificial agents could therefore represent an important intermediate step in the path that leads from narrow artificial intelligence, as exemplified by engines playing chess and Go, to artificial general intelligence (AGI). On the more practical side, there is no doubt that such agents, if indeed able to get by on the financial markets, could be leveraged for highly profitable ventures.

As further evidence of this potential, it should be added that trading is an activity that has now become very common and widespread, and practiced daily by millions of human beings, whose experiences make available concrete and verifiable indications of trading practices. Indeed, according to "The Modern Trader" report, published by Broker Notes, there were nearly 14 million traders worldwide in 2018\footnote{https://brokernotes.co/modern-trader/}. Basically, 1 in around 630 people on the planet had been trading online during that year, with a remarkable surge with respect to the already impressive numbers of 2017, when the online traders totalled worldwide 9,6 million.   In mid-2021, at the time of writing, this growth trend may well have jumped an order of magnitude or so, as a result of the growth of online activities due to the physical isolation imposed by the Covid-19 pandemic. The company Robinhood alone, which through its app strongly contributed to the spreading of trading among the younger generations up to the millennials, counted 18 million accounts as of March 2021\footnote{https://www.cnbc.com/2021/07/01/robinhood-has-18-million-accounts-managing-80-billion-after-rapid-one-year-growth-ipo-filing-shows.html}.

The purpose of this essay is to describe an IT architecture for the design of  artificial trading agents, referred to as \emph{cognitive} by their being endowed with capabilities reproducing those of successful human traders.  On the other hand, it is not within the scope of this article to argue for or against the economic merits of trading. The only thing that we can say is that we work under a "no free lunch" assumption according to which well-performing traders that make money are counterbalanced by others who lose it, and that we aim to include our trading bots into the set of good performers. To this end, what really matters is that trading is becoming an increasingly crowded arena and thus offers indisputable bases for large-scale experimentation.  

Methodologically, a key aspect of our approach is the integration of machine learning and data analytics with heuristics. This is where we expect useful insights that can help improve progress towards AGIs. In fact, compared to narrow AIs such as those dedicated to strategy games, artificial trading agents need to deal with much more complex information flows. Consequently, highly selective strategies for information management are in order, hence the central role of heuristics. This integration of heuristics with learning and data processing could be aimed to other contexts too, for example at advancing physical robots beyond the simple behavioural models currently in use. On the other hand, we will show how adaptive AI models experimented with physical robots, such as evolutionary robotics, could be imported into the domain of trading bots to hone their learning and processing capabilities.

The rest of this essay is structured as follows:\\
\Cref{sec:background} characterizes trading and its sister activity of investing and the challenges they pose to a computational implementation through artificial agents; \cref{sec:heuristics-and-ml} highlights the complementarity of the two methodologies of heuristics and machine learning that are the basis of our computational architecture for Cognitive Trading; \cref{sec:financial-heuristics-toolbox} identifies specific financial heuristics that can provide the basis for a financial heuristics toolbox exploitable for the purposes of Cognitive Trading; \cref{sec:finance-learning} illustrates the role of machine learning at dynamically tuning and optimizing the heuristics above; \cref{sec:cognitive-trading} provides a description of the computational architecture for Cognitive Trading and exemplifies its application in the implementation of the trading engine JATB; \cref{sec:conclusion} concludes the paper.

\section{Investing and Trading}\label{sec:background}
Trading is often contrasted with investing, and for good reasons, in that both activities involve the pursuit of profit in the financial markets, but they do it in different ways. Traders move in and out of financial assets within weeks, days, even minutes, with the aim of short-term profits. What matters to them is which direction the asset will take and how the trader can profit from that. Investors have long-term prospects. They think on scales of months or even years and often hold assets during the ups and downs of the market. To this end, they study the industrial and economic aspects related to the financial assets invested, such as the growth prospects of a company, the political stability of a country, etc., to measure attractiveness and risk of the corresponding assets, namely stocks, bonds, etc. Conversely, traders mainly focus on the technical aspects of ongoing trades, although they have recently been paying close attention to statements and rumours on social media for their potential to turn the markets upside down upon a moment notice, especially if traceable to high-impact business leaders (like Elon Musk, to give a case in point). 

However, the two populations of investors and traders are not completely separate species and there is a continuum that leads from one to the other as in a grayscale. %in which the interval between white and black is filled by shades of grey. 
Let’s start from one end of the spectrum, as far away from trading as possible. This corresponds to \emph{passive investing}, characterized by  investments that track an underlying index or create an asset allocation and stick to it over the long term. Passive investors believe in the Efficient Market Hypothesis \cite{fama} which states that asset prices reflect all available information and are all that matters for the evolution of the market. In other words, an asset is worth exactly its price on the market, no more and no less. As a result, a passive investor will prefer investments that target the market in general rather than individual stocks. Investing in a mutual fund or exchanged traded fund (ETF) that tracks the S\&P 500 index is a typical form of passive investment.  This is because the mutual fund's goal is to simply return what the S\&P 500 returns each year. Passive investing is unambitious but safe, in that it always returns exactly what the market returns, which is why it holds undoubted advantages to the average human investor. It does not need professional managers charging high fees for paying attention to the market and striving to optimize the composition of holdings on a regular basis. As sale and purchase of assets is minimized, consequent tax burdens related to frequent capital gains are also avoided. And it's simple: in fact, it's all about choosing an allocation of reliable long-term securities or some passive mutual funds or ETFs and that's it. What bit of advice such an investor may need can be provided by software programs known as robo-advisors, which can be considered simple, but fit for purpose, forms of narrow AI, being essentially a kind of expert systems, i.e. AI technologies lacking learning capabilities that peeped out in the 1980s. They operate as automated wealth management services that provide investors with investment advice and recommendations. To do this, they collect customer data and conduct analysis on them, allocating funds and optimally balancing the portfolio according to their preferences and expectations \cite{Hayes2019TheAC}.

If this is all right for those prudent savers who want to get the best out of their money without venturing into troubled waters or relying on the pricey advice of financial professionals, it is reasonable to expect that state-of-the-art AI technology can go beyond these basic levels of functionality. To get an idea of how this could happen, let's move further in the spectrum of financial types and take a look at \emph{active investing}. Active investors aim to maximize their returns through the purchase and sale of investments. An  active  investor  continuously  monitors  the performance  of the assets she owns to  take  advantage of  profitable  conditions,  which  may  involve  changing  their allocation  based on  how  the  market  is  going,  for  example by  trying  to  make  short-term  profits  on  the  price  swings  of specific stocks. A more cautious and defensive form of active management is to change the asset allocation to adapt a long-term plan to market instability. For example, an investment portfolio split 60\% in equities and 40\% in bonds enters the risk zone if facing an impending recession, as stocks tend to be hit hard by economic crises. A precautionary move can therefore be reducing equity exposure to 40\% and increasing bond holdings to 60\% to contain losses in the event of a decline in stock values. 

Clearly, active investing is challenging and risky with respect to passive investing, and therefore those who practice it successfully are mostly financial professionals. Furthermore, the more speculative and dynamic it is, the closer it comes to trading, although there is always a litmus test to discern investors, as active and reckless as they may be, from traders. In fact, unlike passive investors, active investors do not believe in the perfect alignment between prices and values of assets assumed in the Efficient Market Hypothesis. For this reason, they will try to do good business with the purchase of under-priced assets by looking at such objective data as patents and other types of intellectual capital of listed companies, industrial and commercial sectors with high growth potential, long-term demands for certain commodities, etc.; and conversely, if on the basis of similar considerations they believe they hold overvalued assets in their portfolio they will liquidate them before they are depreciated by the market. This said, active and passive investors differ in their investment strategies but agree on viewing the market as determined by factors external to the market itself and therefore resting on objective factors. Not so for traders, for whom the market is in fact self-referential, an ecosystem determined by its participants and how they interact with each other. That is why traders rely on both the emotional and analytical propensities of other traders as well as on information imbalances to make gains in much shorter times than those contemplated by even the most dynamic among active investors - times sometimes squeezed down to a handful of seconds. The psychology of the trader appears indeed more complex than that of the investor, who just wants to get good returns from invested capitals, for it includes, beside the desire to make money, the satisfaction deriving from outsmarting others and the quest for the thrill of victory on the pitch. For this reason, trading is evolving into a mass phenomenon, thanks to apps such as the aforementioned Robinhood and to emerging markets such as those of cryptocurrencies, by being not just a way to get rich or impoverished, but also a filler and an existential diversion for millions of individuals whose life takes place more and more in the virtual world rather than in the physical one.

The features that AI must develop to play an effective role in active investing and trading are certainly more demanding than those of passive investing. Detailed and in-depth data acquisition and analysis is a mandatory prerequisite for the activities at the core of active investing, namely identifying and predicting the performance capability of undervalued companies; moreover, this ability to detect underpriced situations must be applicable in a general way so as to systematically pick them out from the market. To this aim data analytics and machine learning techniques are of the essence. In the case of trading, the situation is even more challenging because it is not only a question of spotting out objective elements, not yet reflected in price, that contribute to the value of the negotiated assets, but also of understanding the strategies as well as the emotional propensities of other traders to advantageously ride the sentiment and the mood of the market. The strategic component may suggest that methodologies used for artificial agents that have been successful in strategy games can be straightforwardly applied also to trading, but this is by and large illusory.

It is indeed tempting to view the trading environment as a multiplayer game with imperfect information (as no player has a full knowledge about the full set of resources available to the other players) with multitudes of agents all concurring to the playing. We know that imperfect information multiplayer games, such as poker, have long been within reach of AI methodologies \cite{DBLP:journals/cacm/BowlingBJT17}, but here we are talking about radically different scales, given the number of playing agents who can leave or join the game at any given time. This is in fact an inherently non-stationary environment. Market conditions mutate and agents join and leave, and constantly change their strategies. Can we train agents who learn to automatically adapt to shifting market conditions, without “forgetting” what they have learned before? For example, can an agent successfully transition from a bear market to a bull market and then return to a bear market without needing to be re-trained from scratch? Can an agent learn to identify anonymous traders, who are often involved in large trading operations, and thus be alerted that something big is going on? Can an agent adapt to new types of agents joining the game so that it can live with them, perhaps even to its own advantage? 

Managing these levels of complexity by relying just on learning approaches, even those used successfully for strategy games, appears problematic, precisely because there is a lack of homogeneity, uniformity and even consistency in the information that is supplied to the agent for training purposes. In fact, this appears to be the typical case where complex questions do not admit simple answers, and therefore satisfactory solutions cannot but be themselves, to some extent, complex. However, we can aspire to a principle of structural coherence  that guides us in the building of good solutions and avoids bad designs.  This is the purpose of the next section, where we illustrate how the design of trading agents aimed at satisfactory performance on the financial markets can evolve from the integration of two complementary methodologies. One is given by the nowadays famed and touted machine learning, the other derives from a concept at least equally rooted in the history of artificial intelligence: heuristics.

\section{Heuristics and Machine Learning}\label{sec:heuristics-and-ml}
Heuristics and machine learning have an orthogonal relationship with data: one works selectively, by singling out data strictly relevant to decisions to be made, while the other treats data as a lifeblood that enables the learning of behaviours appropriate to the tasks to be performed. The age of “Big Data” finds both approaches ready and prepared to seize the opportunities and face the challenges: heuristics by making it possible to set course through the tsunami of data towards precise landing points, machine learning by enabling floatation on such a mighty wave by exploiting its power rather than being overwhelmed by it. The two approaches have often been contrasted as alternative views of AI but can actually complement each other effectively. Heuristics can sift through data and give them structure and organization in a simple and efficient way in cases where machine learning would exceed realistically viable requirements for resources and time. For its part, machine learning can refine heuristic strategies whose best execution depends on the setting of parameters to be chosen from a quantity of available data.

Indeed, trading and active investing are typical domains that can benefit from the integration of heuristics and machine learning. In fact, learning to predict market trends and asset performance based on past trends and performance stands as inherently difficult, even with computational resources in abundance. This is due not only to the enormous amount of data to be processed, but also to their dimensional complexity, which, beside price, is determined by such factors as market context (bearish, bullish, stationary, roller coaster), cost of money, level of inflation, unemployment rate, etc. By contrast, proven heuristics obtained from trading practice provide attractive designs for basic artificial traders, whose performance is in turn ameliorable by honing their execution parameters with machine learning - a task made feasible by narrowing it down to heuristically relevant data.

Known and put into practice since ancient Greece, heuristics was successfully proposed and systematized in the second half of the last century by Herbert Simon \cite{Simon1977}  as an effective method of problem solving intended for human decision makers such as managers and investors, as well as for artificial agents. Simon saw heuristics as the natural way to solve problems whenever \emph{bounded rationality} \cite{Simon1990} applies, that is, when the uncertainty and complexity of what is going on does not let an agent make decisions that can be verified as algorithmically optimal; or, put in a nutshell, when an agent needs to cope in one way or another with the roughness of the real world. The pioneering contributions by Simon to the disciplinary fields of both economics and AI are well-known and the concept of bounded rationality was in his view applicable to both contexts, even if in recent decades it has been pursued far more extensively in economics, staying instead rather overlooked in AI (just as AI was substantially unrelated to the developments of the Simon-inspired branch of behavioural economics). Recently, however, the theme of explicitly addressing the boundaries of the rationality criteria applicable in tasks performed by artificial agents is becoming increasingly relevant, even in the context of narrow AI \cite{cassey}. For example, in the games of chess and Go, computers have undoubtedly surpassed humans, but they have not done so by identifying moves that are demonstrably better on strictly formal (logical and mathematical) grounds. The search spaces of these two games are in fact too large to enable an exhaustive, let alone optimal search leading to the identification of the move, or sequence of moves, guaranteed to be the best. Put in another way, this means that neither game is "solved", namely that we do not know what would be the provably best move for any position that may arise. These artificial players instead proceed by trying to achieve positions of the game that appear better on the basis of experiences attributable to games played in the past, so as to devise strategies for coping with limitations of memory and computational capacity that we can interpret to all effects as forms of bounded rationality, albeit rooted in computers rather than in human minds.

The uniformity of the data that characterize strategy games such as chess and GO means that machine learning is in this case sufficient to generate well-performing forms of bounded rationality, as demonstrated by artificial players of superhuman abilities such as the various Alphas (Go, GoZero, Zero) created by the Google company DeepMind \cite{DBLP:journals/nature/SilverSSAHGHBLB17, DBLP:journals/corr/abs-1911-08265}. In the case of trading, the data are much more heterogeneous and the environment in which agents must operate is so intricate that there is need of more complex and articulated strategies that implement ecologically adapted behaviours as well as operate a targeted pre-selection of information. In these strategies heuristics is key. This program matches well with the “fast-and-frugal” school of thought in behavioural economics \cite{Gigerenzer2001}, according to which well-performing behaviours in given contexts all contribute to an adaptive toolbox containing heuristics that can be used according to needs. In the case of trading, the goal is to transfer well-performing heuristics from human practice into artificial trading. These trading agents would be modelled on human cognition, and therefore can be included within Simon's original program of artificial agents that “think” like humans.

Once effective heuristics have been chosen and deployed, machine learning comes into play to refine and consolidate their implementation. Indeed, heuristics applicable in such dynamic and complex contexts as financial markets cannot be completely deterministic, being subject to execution choices that depend on data relevant to the success of trading operations. Here machine learning can produce fundamental improvements, by training heuristics to implement the most appropriate actions with respect to the data that characterize the various market situations. In this regard, and despite the sharing of a common basis in giving a central role to heuristics, our approach differs considerably from the "fast-and-frugal" methodologies. These in fact concern the criteria applied by human individuals who are able to make successful decisions without having to resort to large amounts of data. On the other hand, our goal is to give a direction for the design and implementation of high-performance trading bots which, being machines and therefore unlike humans, will be able to process a lot of data. The intelligence of these artificial agents thus derives from automating heuristics elaborated by the human mind and then augmenting them with the learning abilities already brilliantly demonstrated by the game engines. Following this path, the heuristics toolbox will expand into a full-fledged computational architecture for artificial traders endowed with still bounded yet enhanced rationality. This architecture will also give space to specific tools to get the best out of the emotional and relational information that has flowed into the financial ecosystem since the advent of social media, as well as let trading strategies further evolve through evolutionary meta-heuristics such as genetic algorithms. 

Note that our approach fits well with recent arguments put forward by members of the Deepmind team, which developed the superhuman game engines for chess and Go, according to which all intelligence, from narrow to general, can be explained in terms of reward mechanisms \cite{silversutton}. In fact, given that what matters in trading is to make money, and the more the better, here the association between intelligence and the reward given by the gains made is direct and immediate. However,  rewards are pursued in our framework through a more complex computational architecture than the one based solely on learning that underpins Deepmind's game engines, in that the complexity of trading is best met by the combination of heuristics (namely cumulative knowledge derived from trading practice) and learning (which dynamically adapts heuristics to the needs of the current environment).

\section{A Financial Heuristics Toolbox}\label{sec:financial-heuristics-toolbox}
There are several types of heuristics, more or less known, applicable for trading and investment purposes, including:
\begin{itemize}
\item \textbf{Fast-and-frugal} heuristics, which implement extremely simple yet effective trading methods. One of these, described in \cite{ignorance}, is the idea of investing only in stocks issued by companies that the investor is able to recognize; that is, an investor buys Apple stock only if she knows what Apple is. It is thus shown how this rule of thumb, which selects stocks following the principle that an unrecognized asset is not worth investing, has beaten the market on several occasions. The description of this heuristic in the original article had essentially provocative purposes, demonstrating how sometimes simplicity, if not even simplism, in making decisions pays off compared to more sophisticated and complex approaches. However, it is not senseless to take up its basic inspiration and turn it into a plausible heuristic for an artificial agent. In fact, all it needs is leveraging financial news analysis tools to design the trading bot to always choose company stocks associated with  high  volumes  of  news,  as  they  are,  from  the  bot’s standpoint, highly recognizable; among these it will give preference  to  those  which  are  spoken  of more  positively. Indeed, many instruments providing trading signals based on the analysis of financially relevant content work just like that. Another heuristic, shaped on similar lines of virtuous simplicism, and as a matter of fact vastly put in practice by very simple kinds of trading agents, the so-called high-frequency traders, is given by "grid trading". It boils down to monitoring market trends  so as to sell securities whose value is increasing and buy those that are decreasing. This way profit is made from securities on the rise, while dipping ones are shopped in, under the assumption that they will rise on their turn later on. Recent developments have shown the possibility of boosting grid trading by integrating it with machine learning methodologies, a
confirmation of their complementarity and compatibility with financial heuristics \cite{rundo}.
\item
\textbf{Statistical arbitrage} strategies, on the other hand, are mathematically very sophisticated but ultimately rely on a very simple heuristic principle: the securities held in the portfolio are assigned different scores which are constantly updated based on the market trend, with variations in the score which determine consequent purchase and sale decisions. Statistics plays a central role by finding correlations relevant to decision making. For example, pairs trading \cite{pairs}, one of the most practiced forms of statistical arbitrage, monitors securities that are statistically associated in their trends to exploit periodic deviations in the correlation, as in the following example from Wikipedia\footnote{https://en.wikipedia.org/wiki/Pairs\_trade}: "Pepsi (PEP) and Coca-Cola (KO) are different companies that create a similar product, soda pop. Historically, the two companies have shared similar dips and highs, depending on the soda pop market. If the price of Coca-Cola were to go up a significant amount while Pepsi stayed the same, a pairs trader would buy Pepsi stock and sell Coca-Cola stock, assuming that the two companies would later return to their historical balance point. If the price of Pepsi rose to close that gap in price, the trader would make money on the Pepsi stock, while if the price of Coca-Cola fell, they would make money on having shorted the Coca-Cola stock. The reason for the deviated stock to come back to original value is itself an assumption. It is assumed that the pair will have similar business idea as in the past during the holding period of the stock."  Timely decision making is crucial to the success of statistical arbitrage strategies, so as to avoid missing windows of opportunities to initiate buy/sell actions, which is why they were among the first to be automated, to take advantage of the speed and precision of computerized execution.
\item
\textbf{Micro-heuristics}: with this term we refer to heuristics developed individually by successful traders over years of practice. They are mostly highly idiosyncratic, by depending on a variety of very specific factors, such as the market in which the trader operates, the players with whom he is, metaphorically speaking, used to sitting at the gaming table, the level of risk he is willing to take, etc. For example, a trader may have developed a heuristic, whereby in the first place he looks at the current bookings of operations, checking for the presence of other traders capable of moving large volumes; if this is confirmed, he consequently synchronizes on the big players' activities, trying to opportunistically exploit their behaviour. Transferring these micro-heuristics in the toolbox is practically feasible nowadays, as the so-called copy-trading \cite{DBLP:journals/mansci/ApesteguiaOW20} made possible by numerous platforms allows copying the operations of traders willing to make their way of operating visible in exchange for fees. Once properly documented and regimented, these micro-heuristics can then be automated. 
\item

\textbf{Price action} \cite{Brooks2011}
is a trading methodology applicable on any liquid market and based only on prices and their movements. For this purpose, it analyzes the behavior of prices and their variations by applying a strong component of market contextualization. Price analysis is often done mostly on a daily basis, or possibly over longer timelines such as weekly, or whatever timescale is deemed appropriate. Price action is in a sense a heuristic based on raw prices, whose operation hinges on market trends, by identifying and interpreting its evolution, with particular attention to breakouts and reversals.
 \item
 \textbf{Technical Analysis} \cite{Lo00foundationsof}
builds up on price action in that it leverages past prices in calculations that can then be used to inform trading decisions. It thus groups a set of indicators to make trading choices that sediments generations of practice in a sort of collective memory of how to trade. On the one hand, technical analysis enjoys wide diffusion and adoption, on the other hand it continues to be the subject of numerous criticisms that view it as too general to be truly effective.  Furthermore, a basic objection is that, since these are collectively shared criteria, if everyone applied them no one would gain, as a consequence of a simple application of the “no free lunch” principle. In reality, all objections and contraindications notwithstanding, technical analysis is used extensively by a community that ranges from retail traders to investment banks. Technical analysis lends itself easily to automation because of its highly regimented format in terms of indicators.
 \item
\textbf{Fundamental Analysis} is not in itself a heuristic but provides the context in which to develop good heuristics in the trading and investing sectors by studying in depth the domains in which these activities take place. This is because the business model of a company or an industrial sector, or of currency and commodity markets, their short- and long-term growth prospects, their level of innovation etc are all elements that can contribute to wise choices on how to move in these areas. To these aspects that are the traditional targets for fundamental analysis \cite{graham2009} we must add the study of massively shared digital content, a recent but essential phenomenon as a consequence of the breaking into the trading and finance world of social media such as Facebook, Twitter and other blogs, a phenomenon which particularly concerns "natively digital" securities such as the cryptocurrencies. In fact, these contents convey opinions and emotions that are able to heavily influence market trends. Another relevant development that we can attribute to the methodological evolution of fundamental analysis as following from the technological evolution of financial products is the analysis of the network relationships between securities and the stakeholders who act on them. Again, this fits particularly well such securities as cryptocurrencies which, being Internet-based,  can be naturally characterized in terms of network relationships.
\end{itemize}

\begin{figure*}[!ht]
\centering
\includegraphics[width=.4\textwidth]{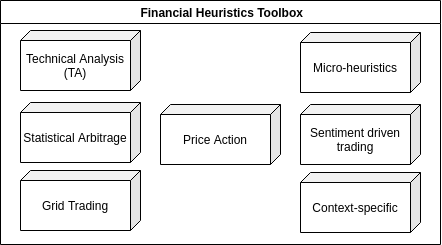}
\caption{Financial Heuristics Toolbox components} \label{fig:financial-heuristics-toolbox}
\end{figure*}

In the following sections we will go into the details of technical analysis and fundamental analysis because they are highly structured methodologies and as such allow illustrating in the most explicit way how heuristics and machine learning can be integrated in order to lay the foundations for a generation of highly performing artificial traders. However, the link between heuristics and machine learning is visible in all the cases illustrated above, grouped together in the Financial Heuristics Toolbox (\Cref{fig:financial-heuristics-toolbox}) as in they all require decisions to be made to optimize profits; when and which ones to take is the aspect that can be refined and made more effective through machine learning.

\subsection{Technical Analysis}
The gist behind technical analysis is that uncertainty reigns supreme in the financial markets due to the huge number of variables involved, and therefore prices and their movements are obviously difficult to predict. Technical analysis deals with observing and studying price movements in order to understand their nature and shed light on their trend, and ultimately to predict future movements. It does so by identifying the patterns that prices form in their fluctuations, following the principle that the succession of prices from moment to moment highlights trends in asset values. Key to this view is the tenet that patterns reflect both market participants and prices. In fact, price follows from the meeting between supply and demand, i.e. it is directly linked to those who own goods and are willing to sell them and those who are willing to buy them. It follows that buyers and bidders usually behave according to rules and thus form patterns.

Well-known and widely used patterns include \emph{head and shoulders}, \emph{rectangle}, \emph{triangle} and \emph{double/triple maximum/minimum}, graphically represented by  subsequent price values on a line. Even more popular are those first identified in 18th century Japan, originally designed to study the rice market and known as \emph {candlesticks}. This terminology stems from their graphical representation through price changes indicated by \emph{Japanese candlesticks} which contain more information than simple line charts. Japanese candles in fact unpack the value of a price over time into four more specific pieces of information: opening price, high price, low price and closing price. The union of these values over a period of time makes up a Japanese candle and, by observing the candles and their conformation, two categories of patterns can be identified: \emph{continuation patterns} and \emph{reversal patterns}. Continuation patterns occur when price changes are constant, i.e. when a trend is still in vogue and therefore continues its course. In contrast, reversal patterns occur when a trend is about to run out of strength and is therefore likely to change its course. Both reversal and continuation patterns are divided into \emph{bullish patterns} and \mkbibemph{bearish patterns}. 

Technical analysis has also a strong statistical component based on the concept of \emph{indicators}, that are mathematical constructions based on statistic models that help to understand movements and their evolution over time. They are divided into five categories:
\begin{itemize}
\item\textbf{Trend Indicators} help to understand which is the dominant trend, i.e. if the trend can be bullish, bearish or sideways. All types of simple/exponential moving averages and their derived indicators belong to this category.
\item\textbf{No Trend Indicators}, also called \emph{oscillators}, are indicators that are not concerned with identifying the trend but with understanding its oscillations and helping to anticipate its reversals. Although \emph{oscillators} are indicators, they are often treated as a separate category given the large number of existing instruments. Some of the most important oscillators are the \emph{Relative Strength Index (RSI)}, the \emph{Momentum} and the \emph{Moving Average Convergence/Divergence (MACD)}.
\item \textbf{Volume Indicators} are concerned with studying the strength of a movement and how it changes over time based on the study of market volumes. Volume refers to the amount traded of an asset and is considered one of the most important aspects to take into account in technical analysis. Some volume indicators are the \emph{On Balance Volume (OBV)}, the \emph{Force Index} and the \emph{Volume Profile Visible Range (VPVR)}.
\item \textbf{Volatility or Range Indicators} are mainly concerned with setting limits to the trend, with high and low limits defining a range in which the price fluctuates. Most popular among such indicators are the \emph{Bollinger Bands} and the \emph{Average True Range (ATR)}.
\item \textbf{Cycle Indicators} are designed to identify cyclical trends of an asset. Markets are subjected to price variations that somehow form cyclical trends and therefore repeatedly make the same movements. Indicators such as \emph{Know Sure Thing (KST)} are designed to detect these cycles.
\end{itemize}

The combination of chart patterns and indicators aims to describe, directly or indirectly, the behavior that most buyers and sellers usually have at a given point in the market. Technical analysis is therefore composed of tools to figure out how markets move based on price changes over time. This results in a heuristic to predict the price trend of an asset which can then be used to automatically carry out market transactions.

\subsection{Fundamental Analysis}
While technical analysis is entirely focused on the subjects that animate the markets, that is the traders themselves, fundamental analysis instead turns its attention to the objects that are traded, that is the assets, in order to assess their risk and potential. It mainly covers the following  three aspects:
\begin{itemize}
\item \emph{Current news}, whether good or bad, gives traders an insight into future market movements and trends, and the sooner they get it the more they can benefit from it, whether it pertains to an industrial sector, or to political upheavals, or to force majeure such as natural disasters.
\item \emph{Macro-economic indicators} are aimed to forecast macro-movements, i.e., the general trends that the world economy is following in a given period. These can be of great help to assess plans to invest in a particular country or continent (i.e. in its national/continental currency).
\item \emph{Market indices} such as the Dow Jones Industrial  Average and the  Dow Jones Transportation Average are meant to provide an understanding of how an industry or a market  is doing and of its  impact on the surrounding economy.
\end{itemize}

\begin{figure*}[!ht]
\centering
\includegraphics[width=.4\textwidth]{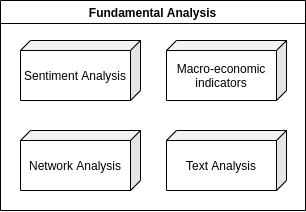}
\caption{Fundamental Analysis components} \label{fig:fundamental-analysis}
\end{figure*}

Macroeconomic indicators and market indices identify well-defined trends that persist for months or even years. For this reason, and also because of taking into account the political context with everything that goes with it, fundamental analysis is of greater interest to active and passive investors rather than to traders who operate on shortened time windows. In fact, whether it is news of a change of government in a country or announcements of companies promising the development of new products, these events produce effects that last over time.\\
\indent However, in its most recent developments fundamental analysis encompasses also financial applications of sentiment analysis \cite{DBLP:journals/entropy/ValenciaGV19, DBLP:journals/es/Wolk20}, as well as of related or complementary disciplines such as content analysis \cite{linton2017dynamic, gorse} and network analysis \cite{diebold2014network,peralta2016network, maesa2018data, giudici2020vector, pichler2021systemic,  rosa}. The purpose of these methodologies is in fact to assess the general mood on negotiable assets before official news is released, or even when there is no official news at all. Mood generally goes hand in hand with the volume of information exchanged, in that when something is talked about a lot it generally means that people start to have strong feelings towards it, either positively or negatively. These mood swings typically translate into market effects within a few days and, being predictable by virtue of the above techniques, can thus be advantageously exploited by traders. Social network phenomena can be similarly measured and taken into account. The full set of methodologies is displayed in \Cref{fig:fundamental-analysis}.

% \subsection{The relation between Fundamental and Technical Analysis}
%Fundamental analysis and technical analysis at first glance seem to be two sides of the same coin, analysing more in detail in reality, we can say instead that technical analysis includes fundamental analysis. Although the fundamental analysis has as its main objective that of studying the macroeconomic aspects of the markets, what gives rise to everything is still the meeting between supply and demand, i.e. the price since technical analysis has as its basis the study of prices in a certain sense this includes fundamental analysis. It is true that although the information in fundamental analysis is contained in price movements, this shows market movements more clearly, but the fact remains that it is still the price that regulates movements.%

\section{Machine Learning for Finance}\label{sec:finance-learning}
Technical analysis helps to understand the movements of assets, by providing tools to do so. But on their own, indicators and oscillators are of little use if they are not employed according to a strategy. %The mistrust that some experienced traders display in their attitude towards technical analysis is explained by the fact that it is a weapon that needs to be handled by expert hands lest it be useless or even harmful. On the other hand, if used to forge good strategies, it can return satisfying results.
Trading strategies consist in combining the various tools provided by technical analysis to identify favorable moments to enter a market and favorable moments to exit it. Once this is done, it is possible to automate both the identification of favorable market points and the execution of entry and exit operations.

The real crux, however, lies precisely in identifying and coding effective strategies, keeping in view it is far from obvious to concoct one that works on all assets and in any market condition. Furthermore, if we intend to transfer strategies from human traders to artificial ones, there could be operations that do not lend themselves to transfer, as it may happen that the trader turns away from technical analysis once in a while and acts instinctively or makes unencodable decisions. Operations that are instinctive or, for whatever reason, cannot be formally modeled, are not transferable to a computer. Nor should it be overlooked that a strategy that has long been profitable may suddenly fail, due to severe market instability. It is therefore paramount to engineer strategies that continuously analyze market conditions and adapt accordingly in order to constantly diversify the operations to be carried out. 

Technical analysis can be usefully compared with machine learning on how to address these issues. Indeed, it is worth articulating in detail the question of whether the two approaches are alternative or rather synergistic, to which we have already anticipated in the previous sections an answer in favor of synergy. First of all, let's consider putting technical analysis aside altogether to rely solely on machine learning which, being data-driven, apparently fits well in an ecosystem as flooded with data as that of trading.

\indent Machine learning is often used to predict the evolution of a phenomenon based on analysis of data obtained by observing the behavior of the phenomenon in the past. Thus, with machine learning it should be possible to predict the evolution of an asset, by analyzing the succession of its price variations. Unfortunately, the price changes alone are not enough to build a suitable learning model for asset evolution. One key aspect for the construction of a good machine learning model is in fact the data set used for training. Building it from the historical price series of an asset amounts to amassing enormous quantities of data, which however are still inadequate for the complexity of the phenomena to be modeled. For example, training a model on low volatility assets does not guarantee its effectiveness on high volatility assets. In addition, market trends and their cyclicality must also be taken into account, which could make ineffective even data sets built out from adequately diversified assets yet associated with specific types of market. Therefore, attention should be paid also to whether data are analyzed on a predominantly bullish, bearish or sideways market. In short, all the variables to be taken into account would impose excessive and unmanageable requirements both for data collection and for computational resources.
% \begin{itemize}
% \item\textbf{Reinforcement learning}
% \item\textbf{Indicator neural network} \cite{dash2016hybrid}
% \item\textbf{Genetic Algorithm} \cite{loginov2021stock} \cite{ozturk2016heuristic}
% \end{itemize}

However, when integrated with technical analysis, machine learning appears to be applicable to trading more effectively and efficiently. In fact, the different choices of  values for the parameters of the indicators of technical analysis consequently generate different strategies that may vary largely in performance. These parameters indicate tolerance thresholds as well as pricing periods, i.e. how many values of the price series to consider for a given calculation. For example, moving averages are determined through a parameter \emph{p} that indicates the previous \emph{p}-values to be considered when calculating the average price at a given time. Human traders use their experience, intuition and technical skills to modify or adapt parameters of this type according to the behavior of an asset. Machine learning could then be used to identify those values that best fit a strategy's indicators and models. In this way, it would be enough to focus the choice of strategy on the search for an effective combination of indicators and patterns, and then improve its performance through machine learning, with the result of drastically downsizing the amount of data to process. This approach appears applicable to technical analysis as well as to the other heuristics that we have described above. 

In conclusion, machine learning is a powerful tool that may well prove effective even in the support of automated trading, yet needs to be driven by a heuristic to cope with the deluge of financial data. A hybrid approach that enables machine learning to take advantage of the indications of a heuristic such as technical analysis could be a win-win in terms of implementation. This same approach could also contribute, for the purposes of the long-term program of evolution of an AGI, a version of bounded rationality driven by both heuristics and data and thus adequately structured to cope with the cognitive demands inherent to the enterprise.

\section{An Integrated Architecture for Cognitive Robotic Trading}\label{sec:cognitive-trading}

Artificial Neural Networks (ANNs) are natural candidates for trading strategy training, by virtue of their many existing successful applications.
However, even a prior selection of the data by virtue of the application of heuristics may not be sufficient to make them, at least in their standard version, computationally feasible in a domain as complex as that of trading \cite{brombinANN}. It is therefore paramount to pick out learning methodologies that are less resource-demanding while being still up to the job. Compared to standard deep learning solutions, an approach that seems to work well here is the combination of genetic algorithms (GAs) and ANNs in a methodology called NeuroEvolution of Augmenting Topologies (NEAT). NEAT is based on the idea of using GAs for the evolution of ANNs. This allows not only to manage the weights on the nodes of a neural network, but also to modify its topology. NEAT algorithms evolve populations of neural networks characterized by similar topologies, where they differ only as regards the internal connections between the nodes, as they hinge on the assumption that it is not necessary to work with completely connected networks and therefore the use of fewer connections leads to less complexity. In fact, much of the waste of resources in standard deep neural network configurations is due to underutilized or poorly performing neuronal connections. A NEAT optimization therefore depends on the ability to find a network with a topology that effectively fits the problem. As shown by \cite{nadkarni2018combining} and \cite{sher2011evolving}, it is an approach that, applied to the trading domain, has so far proved very promising although there is still a lot to test and explore. A similar use of evolutionary approaches integrated with machine learning has already been widely explored in robotics in order to refine and perfect the behavior of robots that move in the physical world \cite{DBLP:journals/alife/HarveyPWQT05}. There the results have certainly been effective and the applicability of the approach to the digital environment of online trading does not therefore come as a surprise. Yet it is at the same time innovative because, compared to hardware robots, trading bots have to face decidedly more complex tasks even if they are exempted from physical risk: it is the difference that there is between knowing how to get by in collecting chemical footprints from rocks on the Martian surface and being able to earn on financial markets that move daily enormous amounts of money by leveraging financial instruments designed according to the highest standards of mathematical sophistication. Furthermore, trading bots as in our architecture can access both learning and heuristic features, as a result of which they can implement much more flexible and versatile behaviors than those so far made available by evolutionary robotics, which refer to the reactive model developed at the end of the 80s, and are therefore completely devoid of heuristic support \cite{DBLP:journals/ai/Brooks91}.

\begin{figure*}[!ht]
\includegraphics[width=\textwidth]{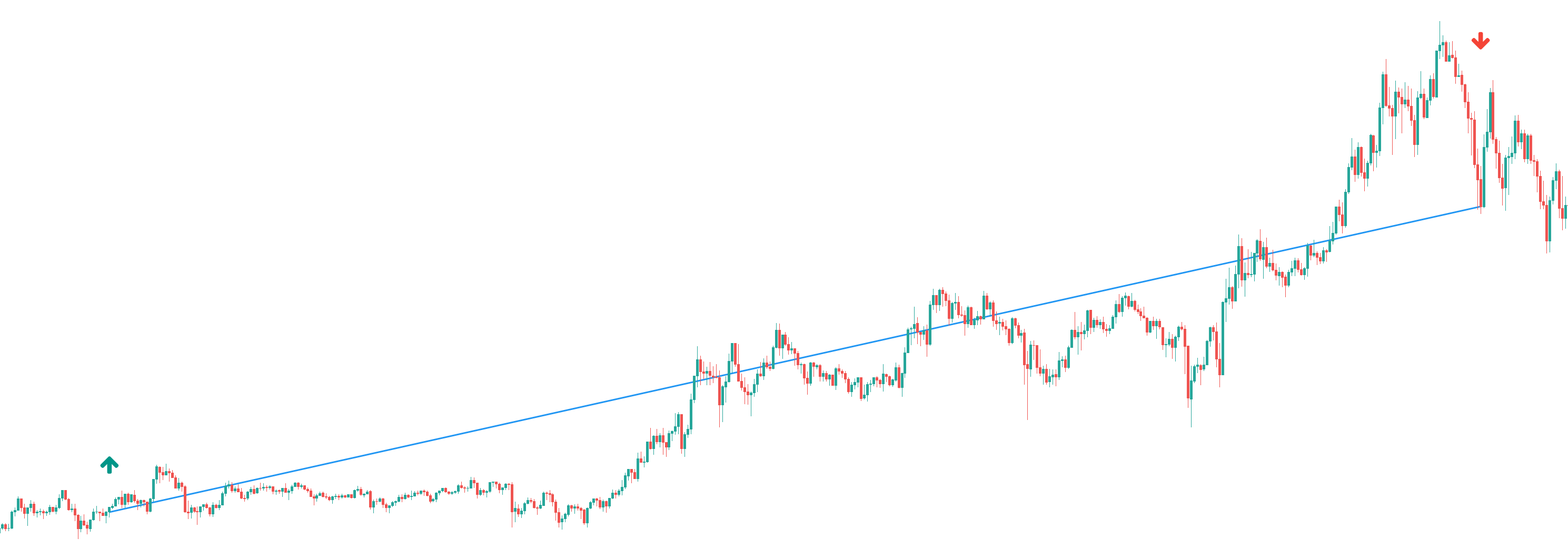}
\caption{Automatic example trade: \emph{ADAUSDT} pair, purchased on April 28, 2021 at an average price of \$1.2983, sold on March 16, 2021 at an average price of \$2.0223 for a profit of approximately 56\%. Used indicators were OBV, RSI, MFI, CCI, WR, ADX and ATR.} \label{fig:automated-trade-example}
\end{figure*}

So let's take a look at how neuro-evolutionary algorithms can boost technical analysis so as to make it more effective and performing and also, in a longer-term perspective, take a relevant step towards cognitive trading. The starting point are the values of the technical analysis indicators as inputs to the algorithm. 
The output of a neuro-evolutionary process applied to technical analysis (\Cref{fig:automated-trade-example}) amounts therefore to the configuration, through a genetic algorithm, of a neural network with a specific topology and weighted arcs. Since the initial layer of the network is composed of technical analysis index values, and the network topology follows from linking together the index values, and, finally, the weights act as thresholds, we can conclude that the end result of applying this methodology is an optimized technical analysis strategy. In short, if we could inspect the network resulting from the neuro-evolutionary algorithm, we would visualize the implemented strategy in the form of the links and weights that make up the network. From a cognitive standpoint, this would amount to peering into the mind of a trading agent, namely of an artificial agent acting within an ecosystem that is far more complex and dynamic than the ones inhabited by the computationally powerful yet cognitively simple engines applied to strategy games. 

\begin{figure*}[!ht]
\centering
\includegraphics[width=\textwidth]{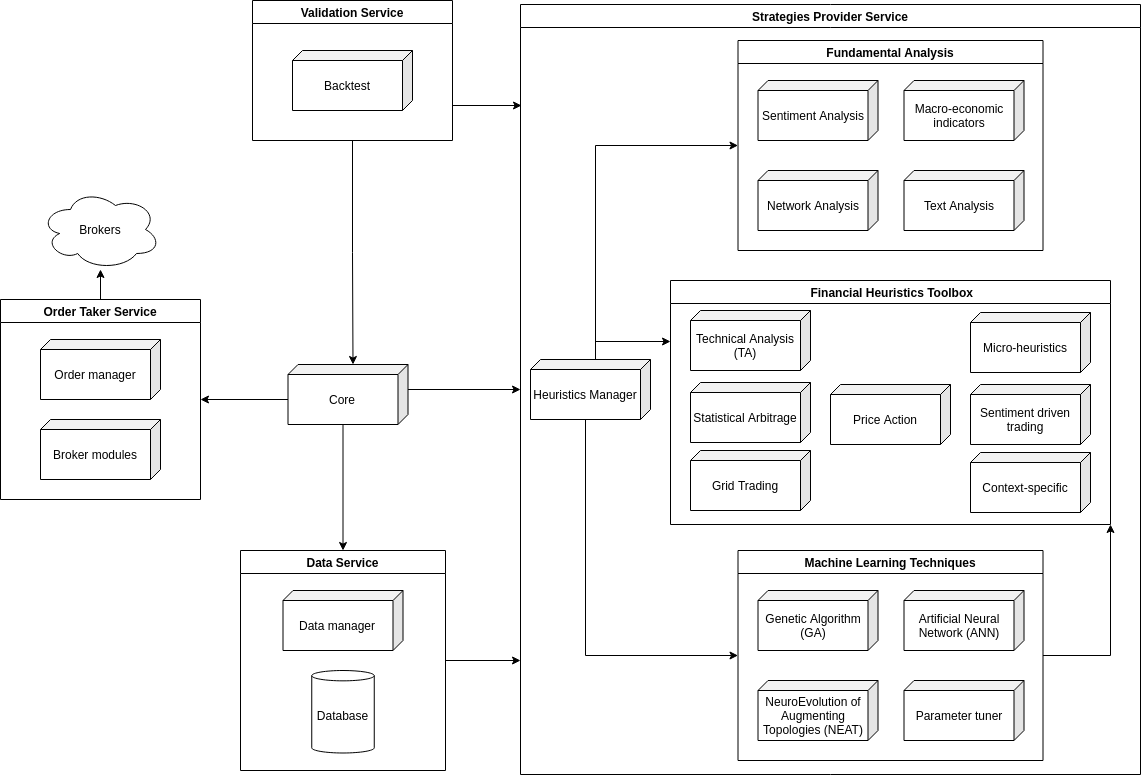}
\caption{Trading system architecture} \label{fig:architecture}
\end{figure*}

We are now able to define a high-level architecture (\Cref{fig:architecture}) that integrates the various components that we have identified and characterized as relevant for an advance towards artificial agents endowed with capabilities of cognitive trading. It is composed of four services and a core module.
\begin{itemize}
\item\textbf{Core (Central module)}, whose purpose is to ensure the interaction between the various components, by handling data management and the correct use of services. The Core mainly deals with initializing the system and using its services to analyze the market and, subsequently, create a strategy and validate it. Orders will then be placed to brokers based on the strategy.

\item\textbf{Order Taker Service}, with the task of interfacing with external systems, namely the brokers, to place orders. This is done by setting up a module/interface for each broker and by enabling the Core to interact with the various brokers through the Order Manager, 

\item\textbf{Data Service}, with the task of collecting market data. This is done by setting up appropriate APIs that enable specific services, such as crawlers and access to external data banks. The collected data is warehoused, as well as constantly updated and made accessible to other services.

\item\textbf{Strategies Provider Service}, which delivers the essential capability of the system to dynamically define and implement trading and 
investment strategies. The Strategies Provider Service is designed precisely to create market-specific strategies. Strategy definition is driven by heuristics, here exemplified, but not restricted to, technical analysis. %, and may well include micro-heuristics developed by individual traders, as mentioned earlier, as well as pairs trading strategies \cite{krauss2017statistical}. 
\begin{itemize}
  \item These heuristics can in turn be input to optimizations based on AI and evolutionary methodologies. Once the heuristic is selected, Genetic Algorithms, Artificial Neural Networks and NeuroEvolution of Augmenting Topologies can enter the stage for the purpose of strategy development.
  \item The Parameter tuner module provides another way to improve the parameters of the strategies, by refining them according to needs. It supports strategy optimization with respect to specific contexts such as a particular type of market or asset.
  \item Finally, the Data Analytics module puts fundamental analysis into the game, by providing tools to conduct market and asset studies. These studies can both provide preliminary support for heuristics and form also the basis for longer-term investment strategies.
\end{itemize}
\item\textbf{Validation Service}, whose role is testing the strategies proposed by the Strategies Provider Service. In particular with the Backtest a strategy can be tested on and scored against historical data contained in the Data Service. Note that the score may vary based on the type of data used for testing, as the same strategy can be applied to different markets, resulting in varying performance.
\end{itemize}

Based on this high-level architecture it was possible to develop an open-source software for automatic trading, dubbed JATB, for Just Another Trading Bot\footnote{https://github.com/ZappaBoy/JATB}, operating as an automatic spot trading system in the cryptocurrencies market and aimed at supporting strategy definition. The cryptocurrency market is one of the most volatile and offers higher profits by trading promptly. Doing it with every change in the market can be difficult for a trader but easy and feasible for an automated system, especially if we consider the fact that it is a market with no closing hours and therefore active 24 hours a day, 7 days a week. The choice of heuristics fell on technical analysis, due to its proven capacity for good returns in the face of relatively easy implementability and limited computational effort.
The end result is a framework designed to simplify the implementation of trading strategies, by fully automating the management of all components, except for the strategical component where it is the trader to have the last say. Therefore, in JATB the trader is relieved from dealing with the operations of the components having only to organize them according to his needs through the configuration options. In addition, JATB is equipped with a backtesting component to let the user run tests to measure the performance of the strategy in a test environment without having to rely on the availability of data from the brokerage platform.
Finally, in order to simplify as much as possible the definition of the strategy, the most used technical indicators, as well as a stop-loss/take-profit dynamic component and a trend identification component are supported. All the components and the parameters of the indicators are fully configurable and adaptable to one's needs.
\Cref{fig:automated-trade-example} and \Cref{fig:params-optimization} give demonstrations of JATB at work.

The trading strategy is therefore at the real core of the system and can be developed through the toolbox. Whether it is a matter of machine learning techniques, technical analysis or sentiment analysis, JATB manages the strategic choices without need of ad-hoc extensions. For its correct functioning in fact it just needs that the strategy provides at each iteration two lists of operations in order to, respectively, open and  close trades, fully delegating the framework with their execution.
JATB is therefore an open-source, highly extensible and configurable framework that enables the trader to define any kind of heuristic she wishes and to automate its execution.

%JATB is already integrated with the API of the broker, with a database already set to save the necessary historical data and it is also executable in multiple instances thanks to the use of the \emph{Docker Engine} that allows the containerization of the system making the deployment of the framework as simple as possible. The backtesting component allows to perform tests to identify the performance of the strategy in a test environment without making open/close requests to the broker. Finally, in order to simplify as much as possible the definition of the strategy, the most used technical indicators, a stop-loss/take-profit dynamic component and a trend identification component are implemented. All the components and the parameters of the indicators are fully configurable and adaptable to one's needs.\\
%The trading strategy is therefore the real core of the system and it can be developed through heuristics. Whether it is a matter of machine learning techniques, technical analysis or sentiment analysis, JATB manages the strategic choices without the need of particular integrations. For the correct functioning in fact it is only necessary that the strategy provides at each iteration a list of operations to open and one of operations to close delegating to the framework the complete execution of the operations necessary for the correct functioning.
%JATB is therefore an open-source, highly extensible and configurable framework that allows a trader to define any kind of heuristic he wishes and to automate its execution.

\section{Conclusion}\label{sec:conclusion}
In this article we have pursued the construction of a high-level architecture for the design and implementation of artificial trading agents, with the idea of achieving two objectives: a more practical and immediate one, which contributes to the progress of the state of the art of trading bots, and a longer-term one, as part of the research program aiming at artificial general intelligence (AGI). With respect to this second objective, we have characterized the degree of progress thus achievable as an intermediate step between narrow AI, as represented by programs capable of playing at superhuman levels in strategy games such as chess and go, and AGI.  This is because the trading agents’ theater of operation is much more open and complex than that of the agents who operate on strategy games, but on the other hand they still fall short of general intelligence, whether natural or artificial. We refer to this intermediate form of intelligence as Cognitive Trading, two terms that aptly capture both the advancement, with respect to narrow AI, resulting from the cognitive ability to make choices and decisions in a complex environment, and the limitation inherent in the specificity of the application domain.%, although we expect to see them evolve to reach and even surpass the skill of the best human traders.

In defining the conditions for achieving the second of these objectives, we indicated how, unlike narrow AI, the complexity of the application domain here prevents relying on a single technology. To this end, we introduced the concept of toolbox, importing it from the cognitive approach in behavioral economics, and expanding it beyond the use of heuristics practiced there with data-driven tools such as, first of all, machine learning, which underlies the recent successes in narrow AI. Since the systematization of heuristics itself comes from artificial intelligence, our approach thus reconciles two different traditions from the same disciplinary field, something that comes as very natural in the context of trading and investing, since these sectors are rich in techniques and methodologies which by all means correspond to heuristics that were widely tested through years of adoption by human traders.
Further extensions of this expanded toolbox come from  contextualizing trading and investing activities in social media, with tools such as sentiment analysis, content analysis and social network analysis. This has led to the articulation of a full-fledged computational architecture to support Cognitive Trading, that can be thought as accessible and customizable according to a wide choice of options and their combinations, assuming adequate constraints on their order and pairing; relevant indications are that heuristics should be applied before machine learning, rather than vice versa, and that the latter can be optimized by pairing with evolutionary methodologies such as genetic algorithms. %Going for machine learning implies appropriate computational resources, and that's another consideration to take in account. There is therefore something for all budgets, and it all comes with both with precautions and with appropriate optimizing indications.

This flexibility and modularity makes the first objective immediately achievable, namely  to undertake at once the development of practical implementations derived from the architecture, by leveraging modern software development practices predisposed for the incremental addition of functionalities; as demonstrated by the aforementioned JATB bot, which currently supports heuristics in the form of technical analysis and is in the course to be integrated with machine learning and content and sentiment analysis capabilities.

\begin{figure*}[!ht]
\caption{The following graphs show what can be the results of optimizing indicator parameters, in this case the EMA indicator. Exponential Moving Averages (EMA) are a type of Moving Averages (MA) that give more importance to more recent values over time. The EMA indicator is mostly used for trend identification and often a cross between two EMAs of different period is used as a trend reversal signal. The period indicates how many price values are taken into account for the calculation of the indicator, i.e. the parameter we are interested in optimizing. In the following candlestick charts two EMAs are shown, one with a shorter period (slow EMA - black line) and one with a longer period (fast EMA - orange line), the signals of purchase (blue circles) and sale (purple circles) being given by their intersection. It can therefore be seen that initially there are several false sell signals according to the trend continuation (\Cref{fig:param-optimization-9-21}). With the optimization of the choice of the parameters the false signals diminish sensibly (\Cref{fig:param-optimization-9-30}, \Cref{fig:param-optimization-20-30}) until arriving to an optimal solution (\Cref{fig:param-optimization-20-50}).} \label{fig:params-optimization}
\begin{subfigure}[b]{\textwidth}
    \centering
    \includegraphics[width=\textwidth]{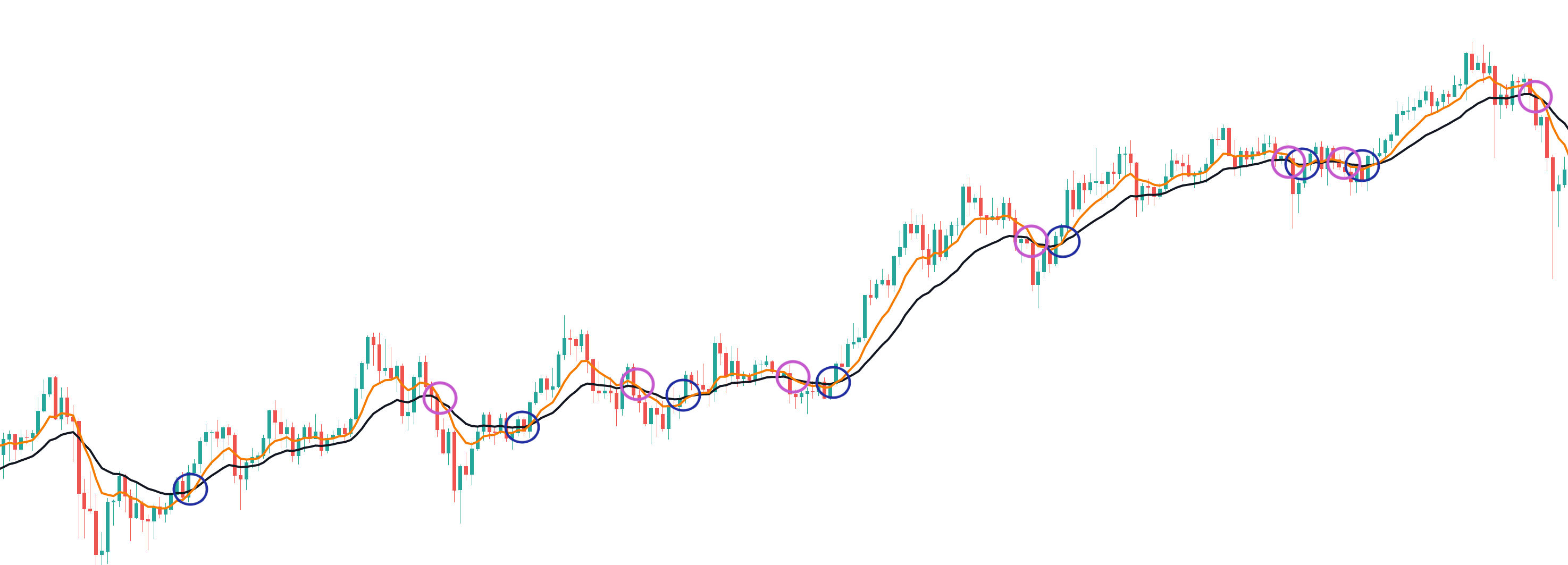}
    \caption{Slow EMA: 9, Fast EMA: 21}
    \label{fig:param-optimization-9-21}
\end{subfigure}

\begin{subfigure}[b]{\textwidth}
    \centering
    \includegraphics[width=\textwidth]{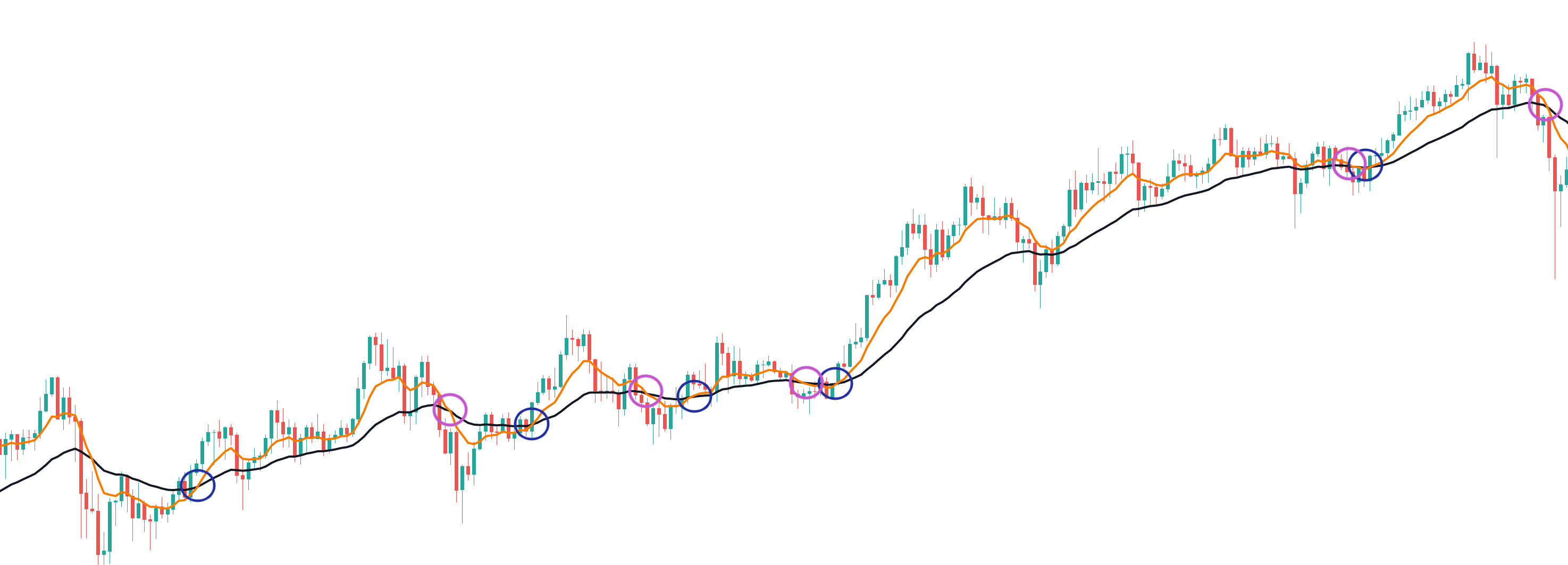}
    \caption{Slow EMA: 9, Fast EMA: 30}
    \label{fig:param-optimization-9-30}
\end{subfigure}
\end{figure*}

\begin{figure*}[!ht]\ContinuedFloat
\begin{subfigure}[b]{\textwidth}
    \centering
    \includegraphics[width=\textwidth]{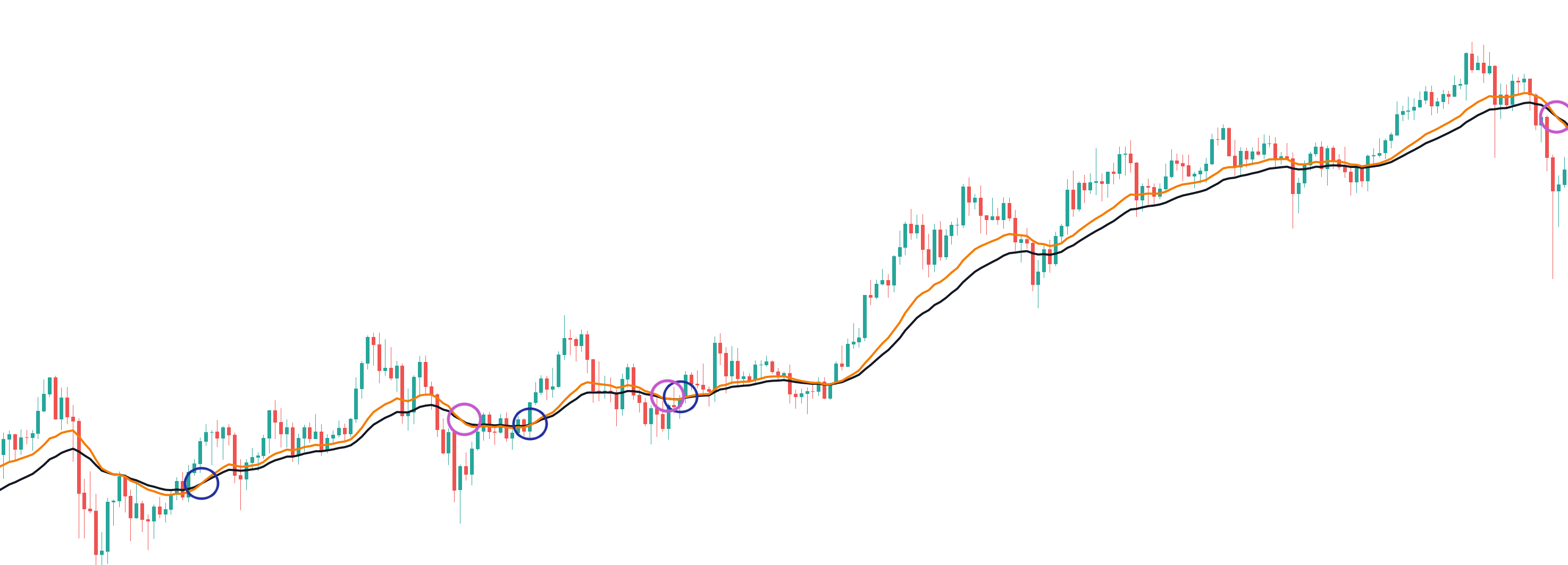}
    \caption{Slow EMA: 20, Fast EMA: 30}
    \label{fig:param-optimization-20-30}
\end{subfigure}

\begin{subfigure}[b]{\textwidth}
    \centering
    \includegraphics[width=\textwidth]{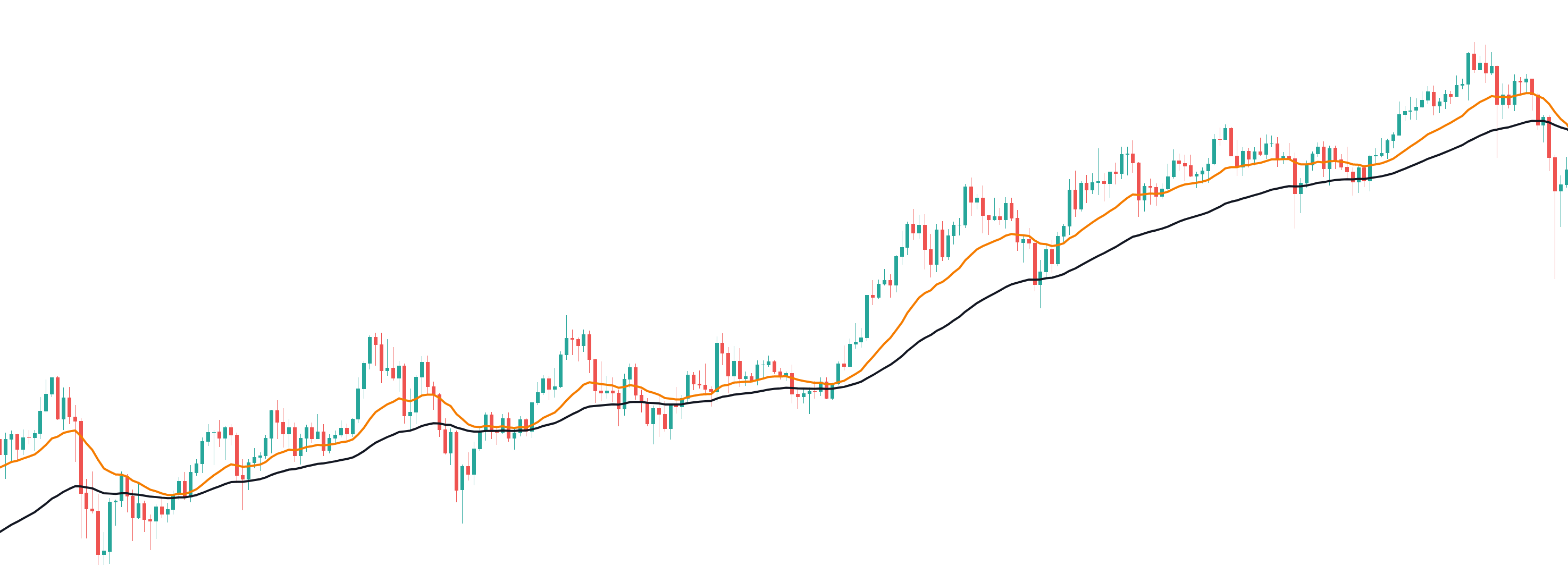}
    \caption{Slow EMA: 20, Fast EMA: 50}
    \label{fig:param-optimization-20-50}
\end{subfigure}
\end{figure*}

\clearpage

\printbibliography

\end{document}